\documentclass{article}
\usepackage{perpage} 
\MakePerPage{footnote} 
\usepackage{ijcai22}

\usepackage{times}
\usepackage{soul}
\usepackage{url}
\usepackage[hidelinks]{hyperref}
\usepackage[utf8]{inputenc}
\usepackage[small]{caption}
\usepackage{graphicx}
\usepackage{amsmath}
\usepackage{amsthm}
\usepackage{booktabs}
\usepackage{algorithm}
\usepackage{booktabs}
\usepackage{multirow}
\usepackage{subfigure}
\usepackage{physics}

\usepackage{newfloat}
\usepackage{siunitx}
\usepackage{listings}

 \usepackage{amsfonts,amssymb}
\usepackage{bm}

 \usepackage{amsfonts,amssymb}
\usepackage{algorithmicx}
\usepackage{algpseudocode}
\usepackage{bm}

\title{Rethinking the Promotion Brought by Contrastive Learning \\to Semi-Supervised Node Classification}

\author{
    Deli Chen\textsuperscript{\rm 1}, 
    Yankai Lin\textsuperscript{\rm 1}, 
    Lei Li\textsuperscript{\rm 2}, 
    Xuancheng Ren\textsuperscript{\rm 2}, 
    Peng Li\textsuperscript{\rm 3},
    Jie Zhou\textsuperscript{\rm 1}, 
    Xu Sun\textsuperscript{\rm 2}
    \affiliations
    {\small
    \textsuperscript{\rm 1}{Pattern Recognition Center, WeChat AI, Tencent Inc., China}\\
    \textsuperscript{\rm 2}{MOE Key Lab of Computational Linguistics, School of Computer Science, Peking University, China}\\
    \textsuperscript{\rm 3}{Institute for AI Industry Research (AIR), Tsinghua University, China}
    }
    \emails
    {\footnotesize
    \texttt{\{delichen,yankailin,withtomzhou\}@tencent.com, lilei@stu.pku.edu.cn}\\
    \texttt{\{renxc,xusun\}@pku.edu.cn, lipeng@air.tsinghua.edu.cn}
    }
}

\begin{document}

\maketitle

\begin{abstract}
  Graph Contrastive Learning (GCL) has proven highly effective in promoting the performance of Semi-Supervised Node Classification (SSNC). However, existing GCL methods are generally transferred from other fields like CV or NLP, whose underlying working mechanism remains under-explored. In this work, we first deeply probe the working mechanism of GCL in SSNC, and find that the promotion brought by GCL is severely unevenly distributed: \textit{the improvement mainly comes from subgraphs with less annotated information}, which is fundamentally different from contrastive learning in other fields. However, existing GCL methods generally ignore this uneven distribution of annotated information and apply GCL evenly to the whole graph. To remedy this issue and further improve GCL in SSNC, we propose the Topology InFormation gain-Aware Graph Contrastive Learning (TIFA-GCL) framework that considers the annotated information distribution across graph in GCL. Extensive experiments on six benchmark graph datasets, including the enormous OGB-Products graph, show that TIFA-GCL can bring a larger improvement than existing GCL methods in both transductive and inductive settings. Further experiments demonstrate the generalizability and interpretability of TIFA-GCL.\footnote{Part of the work was done while Peng Li was working at Tencent.}
\end{abstract}

\begin{figure*}[t]
\centering
\includegraphics[width=\linewidth]{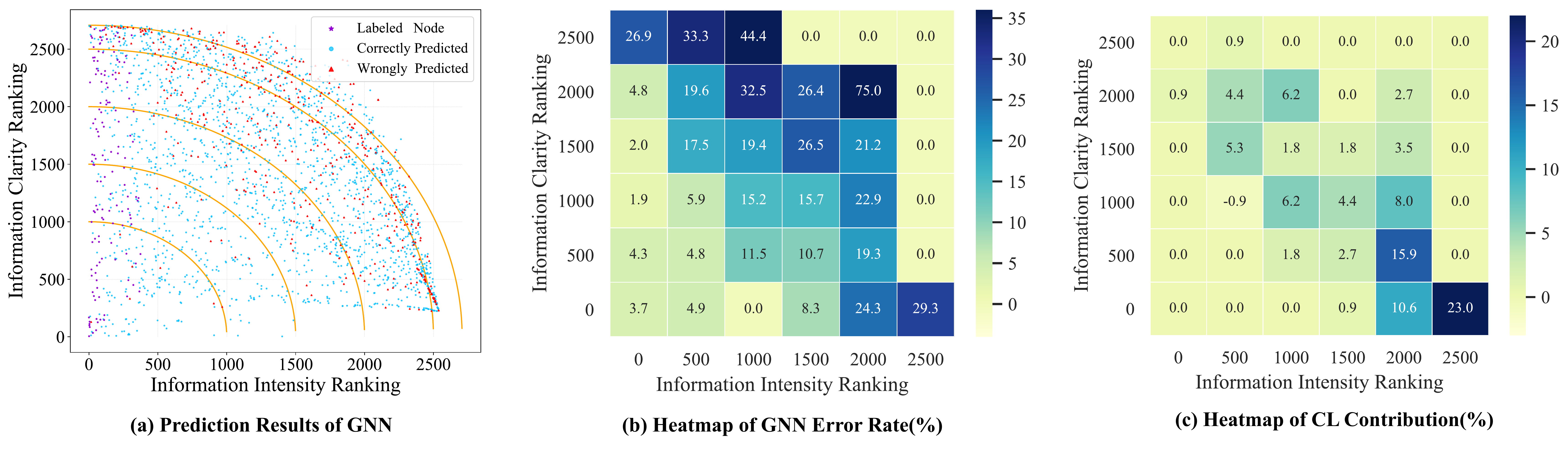}
\caption{
The \textit{RADIAL DECAY} phenomenon in semi-supervised node classification task. \textit{Information Intensity Ranking} and \textit{Information Clarity Ranking} are the node ranking orders of annotated information intensity (Eq.~\ref{equation:info&noise1}) and clarity (Eq.~\ref{equation:info&noise2}) from the highest to lowest, respectively. Subfigure (a) displays the prediction result of GCN model on \textit{CORA} dataset. We can find that graph nodes in SSNC are characterized by a fan-shaped distribution and the GNN's inference ability decays radially outward. 
Subfigure (b) displays the error rate for different grids in subfigure (a).
Subfigure (c) displays the contribution of GCL (gap with GNN) among different grids. We can find that the GCL effect increases radially outward, which is contrary to GNN's learning ability.
}
\label{figure_analysis}
\end{figure*}

\section{Introduction}
Graph Neural Networks (\textbf{GNNs}) can effectively utilize the relation of nodes from graph and 
have shown powerful modeling capabilities~\cite{semi-supervised} in Semi-Supervised Node Classification (\textbf{SSNC}). 
Nevertheless, vanilla GNN training~\cite{model_gcn} accesses the training signal mainly from the limited labeled nodes and pays little attention to the massive information contained in the unlabeled nodes.
To fix this issue, recent studies~\cite{gcl_multiview,gcl_grace,gcl_adaAug,gcl_nipsaug,gcl_cg3,gcl_random_gnn} introduce Contrastive Learning (\textbf{CL}) into graph scene to leverage the information of the unlabeled nodes and have achieved promising performance. 

Though successful, the working mechanism of Graph Contrastive Learning (\textbf{GCL}) in SSNC remains under-explored because the graph scene is fundamentally different from the other scenes like image or text. The samples are connected by complex edges and the task-related annotated information are propagated from the labeled nodes to the unlabeled nodes through graph topology, which results in an imbalanced distribution of the annotated information distribution across the graph.
Therefore, how this imbalanced distribution can affect GCL running and why CL works in the SSNC task have become essential research questions in this field. 

In this work, we first investigate how the annotated information distribution affects the GCL effect. We employ the well-known Label Propagation~\cite{LP3} method to measure the annotated information distribution. We find that \textbf{the promotion of GCL mainly comes from those nodes with less annotated information}. 
Specifically, in Figure~\ref{figure_analysis}, we notice that {GNN's inference ability decays radially outward in the annotated information intensity-clarity coordinate system, while GCL's effect, in contrast, increases radially outward.} However, existing GCL methods overlook this phenomena and apply CL to the whole graph evenly.

Motivated by our findings, we propose the \textbf{T}opology \textbf{I}n\textbf{F}ormation gain--\textbf{A}ware \textbf{G}raph \textbf{C}ontrastive \textbf{L}earning (\textbf{TIFA-GCL}) framework to enhance the GCL methods from two aspects:
(1) from the global view of the whole graph, we enhance the GCL effect on nodes that receive less annotated information from the information source (labeled nodes), by devising novel self-adaptive graph corruption and CL loss weight schedule mechanisms; 
and (2) from the local view of each node, we observe that nodes with similar annotated information distribution are more likely to share the same class label, based on which we promote the contrastive pair sampling and subgraph sampling strategies.

Extensive experiments on six benchmark datasets, including the enormous \textit{OGB-Products} graph~\cite{OGB}, have shown that our TIFA-GCL method can bring more significant improvements to GNN training compared to existing GCL methods in both transductive and inductive setting. 
Further investigations are provided to answer several important research questions of our work.

\section{A Closer Look at GCL in Semi-Supervised Node Classification}
\label{sec:inspection}
Owing to the success of GCL in SSNC, we attempt to probe its working mechanism in this section.
Since nodes' received annotated information has a crucial impact on GNN's inference performance, we first introduce our measurement for the node's received annotated information from the labeled nodes along graph topology, and then investigate how the annotated information will affect the performance of GCL from the perspectives of information intensity and clarity.

\subsection{Task Formalization}
We follow the benchmark setting ~\cite{model_gcn,semi-supervised,model_sage} to discuss the working mechanism of GCL in the semi-supervised node classification task.
Formally, given an undirected graph $\mathcal{G} = (\mathcal{\bm{V}}, \mathcal{\bm{E}})$, where $\mathcal{\bm{V}}$ is the node set with an embedding matrix $\bm{X}\in \mathbb{R}^{n \times h}$ ($n = |\mathcal{\bm{V}}|$ , $h$ is the feature dimension size) and $\mathcal{\bm{E}}$ is the edge set represented by an adjacency matrix $\bm{A}\in \mathbb{R}^{n \times n}$,
the node classification task aims to learn a classifier $\mathcal{F}$ to predict the class label $\bm{y}$ for each node. We denote the labeled and unlabeled node sets as $\bm{{L}}$ and $\bm{{U}}$, respectively.

\subsection{Measuring the Annotated Information from the Labeled Nodes}
In this work, we employ the Label Propagation~\cite{LP3} (\textbf{LP}) algorithm to measure the annotated information distribution among the graph nodes that have complex connections with graph edges. 
LP is a very popular method for SSNC based on the message-passing mechanism: the labels are propagated from the labeled nodes to the unlabeled nodes through graph edges. We adopt LP to measure the annotated information distribution for two reasons: (1) LP focuses on the graph topology, which can reflect the annotated information clearly; (2) LP shares the same mechanism (message-passing) with many GNN models~\cite{LP4}, so the conclusions of it can be transferred to general GNN models like GCN~\cite{model_gcn} and GraphSAGE~\cite{model_sage}. The LP method can be formulated as:

\begin{equation}
     \bm{Z} = \alpha (\bm{I} - (1-\alpha)\bm{A'})^{-1}\bm{Z}^{0},
\end{equation}
$\bm{I}$ is the identity matrix; $\alpha \in (0,1]$ represents the random walk restart probability. $\bm{A'} = \bm{A}\bm{D^{-1}}$ and $\bm{D}$ is the node degree matrix. $\bm{Z} \in \mathbb{R}^{n \ast k}$ is the convergence result of LP; $\bm{Z}_{i,j}$ indicates the probability that the the $i$-th node is belong to the $j$-th class and we regard it as the influence received by the $i$-th node from the $j$-th class's labeled nodes; $\bm{Z}^{0}$ in the initial label matrix where labeled nodes are represented by one-hot vectors and unlabeled nodes are filled with zero. In practice, we take a updated version $\bm{Z}^{\ast}$ to consider the category preference of nodes (shown in Appendix A1).

\subsection{Probing the Mechanism of GCL}
In this part, we analyze the working mechanism of GCL with the help of LP.
Ideally, the received information for each node should be concentrated on one category, which means that the annotated information is strong and clear. 
So we regard the maximum item of vector $\bm{Z}_{i,:}^{\ast}$ to be the information intensity and the gap to the others as the information clarity for node $i$:
\begin{flalign}
\label{equation:info&noise1}
\bm{Q}_i^\text{Intensity} &= \mathrm{max}(\bm{Z}_{i,:}^{\ast}) \\
\label{equation:info&noise2}
\bm{Q}_i^\text{Clarity} &=   \mathrm{max}(\bm{Z}_{i,:}^{\ast}) - \sum\bm{Z}_{i,:}^{\ast}
\end{flalign}
where $\mathrm{max}(\cdot)$ returns the max value of the input vector. 
With $\bm{Q}^\text{Intensity}$ and $\bm{Q}^\text{Clarity}$, we conduct an analysis experiment on the benchmark \textit{CORA}~\cite{dataset_ccp} dataset trained by GCN~\cite{model_gcn} in Figure~\ref{figure_analysis}. 
We plot the prediction results  in the $\bm{Q}^\text{Intensity} \times \bm{Q}^\text{Clarity}$ rectangular coordinate system in Figure~\ref{figure_analysis}~(a).
Figure~\ref{figure_analysis}~(b) shows the error rates for different grids from Figure~\ref{figure_analysis}~(a).
From Figure~\ref{figure_analysis}~(a)(b),
we can find that, the graph nodes in SSNC are distributed in a fan shape, which means that there is no nodes with low $\bm{Q}^\text{Intensity}$ and low $\bm{Q}^\text{Clarity}$ at the same time.
Moreover, we notice that the \textbf{GNN's inference ability has the characteristic of radial decay}\footnote{
This phenomenon can also be found in other datasets. We provide the results in Appendix A2.}:
GNN can predict accurately for nodes around the origin of coordinate, which have both high $\bm{Q}^\text{Intensity}$ and high $\bm{Q}^\text{Clarity}$;
however, the GNN inference ability decays radially outward from the origin of coordinate, where nodes receive more noise (information from other classes' labeled nodes) and less helpful information.
Figure~\ref{figure_analysis}~(c) displays the contribution of GCL effect (the improvement to GNN) among all the grids from Figure~\ref{figure_analysis} (a). 
We can find that just in contrary to GNN's inference ability, \textbf{GCL's effect has the characteristic of radial increase}: the farther the sector away from the origin of the coordinate, the greater the improvement brought by GCL.
Besides, comparing to nodes with high $\bm{Q}^\text{Intensity}$ and low $\bm{Q}^\text{Clarity}$ (upper left in plot), GCL enhances GNN's learning ability for nodes with low $\bm{Q}^\text{Intensity}$ and high $\bm{Q}^\text{Clarity}$ (low right in plot) more effectively.
We analyze the reason lies in that nodes with high $\bm{Q}^\text{Intensity}$ and low $\bm{Q}^\text{Clarity}$ are strongly influenced by graph learning and can be hardly corrected by CL.

Inspired by the fan-shaped distribution of graph nodes and the radical decay phenomenon in GNN, we quantitatively measure each node's Topology Information Gain (short as \textbf{TIG}, indicating the quality of nodes' received annotated information in this work) by jointly considering the node information intensity and clarity. 
The TIG value $\bm{T}_i$ for the $i$-th node is calculated in the following equation:

\begin{flalign}
\label{equation:tig}
\bm{T}_i = \bm{Q}_i^\text{Intensity} + \frac{\lambda}{k-1} \ast \bm{Q}_i^\text{Clarity},
\end{flalign}
where $\lambda$ is the weight of information clarity.  
The GCL effect across TIG bins is shown in Appendix A3.
We summary that: \textbf{The promotion of GCL mainly comes from nodes with less annotated information in the SSNC task.}

\section{Methodology}
\label{section_method}

In this section, we first introduce the whole framework of our method in Section~\ref{section_TIFA-GCL}, and then introduce how we modify the CL loss accumulation (Section~\ref{section_schedule}), graph perturbation (Section~\ref{section_graph_perb}) and contrastive pair sampling (Section~\ref{section_sampling}) to consider the annotated information distribution and promote the CL effect. Moreover, a novel subgraph sampling (shown in Appendix B2) method called SAINT-TIFA is proposed for the inductive setting.

\subsection{TIFA-GCL Framework}
\label{section_TIFA-GCL}
Generally, in semi-supervised GCL, the training objective consists of two parts: the supervised cross-entropy loss and the unsupervised contrastive loss.
Given a GNN encoder $\mathcal{F}$ (e.g., GCN), the supervised cross-entropy loss $\mathcal{L}_{CE}$ is calculated on the labeled set $\bm{{L}}$:
\begin{flalign}
\bm{g} &= \mathcal{F}(\bm{X},\bm{A}, \bm{\theta}), \\
\mathcal{L}_{CE} &= -\frac{1}{|\bm{{L}}|} \sum_{i \in \bm{{L}}}\sum_{c=0}^{k-1} \bm{y}^c_i\,\log\,\,{\bm{g}}^c_i, \label{l1} 
\end{flalign}
where $\bm{g}_i$ is the GNN output for node $i$; $\bm{y}_i$ is the gold label in one-hot embedding; $\bm{\theta}$ is the parameters of $\mathcal{F}$.
Apart from $\mathcal{L}_{CE}$, the contrastive loss in
TIFA-GCL contains two parts: the self-consistency loss and the pair-wise loss with positive and negative pairs.
The self-consistency loss is calculated in the following equation:
\begin{flalign}
\mathcal{L}^i_{s} &= \mathrm{KL}(\mathcal{F}(\bm{X_p},\bm{A_p},\bm{{\theta}})_i,\mathcal{F}(\bm{X},\bm{A},\bm{\widetilde{\theta}})_i),
\end{flalign}
where $\mathrm{KL(\cdot,\cdot)}$ is the Kullback-Leibler divergence function; the augmented node embedding matrix $\bm{X_p}$ and the adjacency matrix $\bm{A_p}$ are generated by the TIFA-graph perturbation which is detailed in Section~\ref{section_graph_perb}. 
Following~\cite{cl_uda} and~\cite{gcl_nodeAug}, $\bm{\widetilde{\theta}}$ is a fixed copy of the current parameter $\bm{{\theta}}$ and the gradient is not propagated through $\bm{\widetilde{\theta}}$.
Similarly, the contrastive loss with positive and negative pairs are computed:
\begin{flalign}
\mathcal{L}^i_{p} = &\frac{1}{|\bm{P}_i|} \sum_{j\in\bm{P}_i} \mathrm{KL}(\mathcal{F}(\bm{X_p},\bm{A_p},\bm{{\theta}})_i,\mathcal{F}(\bm{X},\bm{A},\bm{\widetilde{\theta}})_j)  \nonumber\\ -\, \mu_1 &\frac{1}{|\bm{N}_i|} \sum_{j\in\bm{N}_i} \mathrm{KL}(\mathcal{F}(\bm{X_p},\bm{A_p},\bm{{\theta}})_i,\mathcal{F}(\bm{X},\bm{A},\bm{\widetilde{\theta}})_j),
\label{equation::loss_with_pairs}
\end{flalign}
where $\bm{P}_i$ and $\bm{N}_i$ are the positive and negative pairs for the $i$-th node generated by our TIFA-contrastive pair sampling strategy introduced in Section~\ref{section_sampling}, $\mu_1$ is a hyper-parameters to control the weight of GCL loss of negative pairs comparing to positive pairs.
Then the complete unsupervised loss on node $i$ is calculated via:
\begin{equation}
     \mathcal{L}_{U}^i= \mathcal{L}^i_{s} +  \mu_2 \mathcal{L}_{p}^i,
     \label{equation::self-supervised loss}
\end{equation}
where $\mu_2$ controls the weight of $\mathcal{L}_{p}^i$ relative to $\mathcal{L}_{s}^i$.

\subsection{TIFA-Contrastive Loss Weight Schedule}
\label{section_schedule}
Different from all the existing GCL studies~\cite{gcl_grace,gcl_cg3,gcl_nodeAug,gcl_adaAug,gcl_multiview,gcl_nipsaug,gcl_random_gnn}, which accumulate the unsupervised contrastive loss from different nodes directly, we instead propose to adjust the CL loss weight adaptively among different subgraphs according to their TIG.
Specifically, we set the CL weight $\bm{w}_i$ of each node differently using a cosine annealing schedule based on the node TIG ranking\footnote{We also try other schedule methods like linear schedule, piece-wise schedule and scheduling based on the absolute TIG value instead of the relative ranking. The experiment results show the cosine method works best.}:
\begin{flalign}
\bm{w}_i = w_{\text{min}} + \frac{1}{2}(w_{\text{max}}-w_{\text{min}})(1+\mathrm{cos}(\frac{\mathrm{Rank}(\bm{T}_i)}{n}\pi)),
\label{equation::wi}
\end{flalign}
where $w_{\text{min}},w_{\text{max}}$ are the hyper-parameters indicating the minimum and the maximum values of CL loss, respectively; ${\mathrm{Rank}}(\bm{T}_i)$ is the ranking order of $\bm{T}_i$ from the smallest to the largest. Then the final training objective of TIFA-GCL consists of the supervised loss and a weighted-sum of the unsupervised contrastive loss:
\begin{equation}
    \mathcal{L} = \mathcal{L}_{CE} + \frac{1}{n} \sum\nolimits_{i=0}^{n-1} \bm{w}_i \mathcal{L}^i_U,
    \label{eq_cl_schedule}
\end{equation}
With this manner, subgraphs receiving less annotated information from graph topology can benefit more from GCL, forming a more adaptive combination of graph representation learning and graph contrastive learning.

\subsection{TIFA-Graph Perturbation}
\label{section_graph_perb}
Unlike perturbation in images~\cite{cl_moco1,cl_fixmatch} where samples can be perturbed independently, 
graph nodes interact with each other and the disturbance of one node will spread to its adjacent neighbors. Therefore, if we perturb too many nodes or edges in graph, the augmented graph will be too different from the original one and thus weakening the GCL effect.
We propose that the limited graph perturbation should be more likely to take place on subgraphs with insufficient information gain from graph learning.
Thus, we propose to sample nodes to disturb according to their TIG values: 
a node will be selected with a higher probability for augmentation when it has a lower TIG value. 
Comparing to the contrastive loss weight schedule which directly adjusts the CL weight, weighted graph perturbation is a more essential method to enhance GCL effect on nodes with fewer TIG values since graph perturbation is the upstream operation of computing CL target. The details of TIFA-graph perturbation algorithm is shown in Appendix B1.

\begin{figure}[t]
\center
\includegraphics[width=0.95\columnwidth]{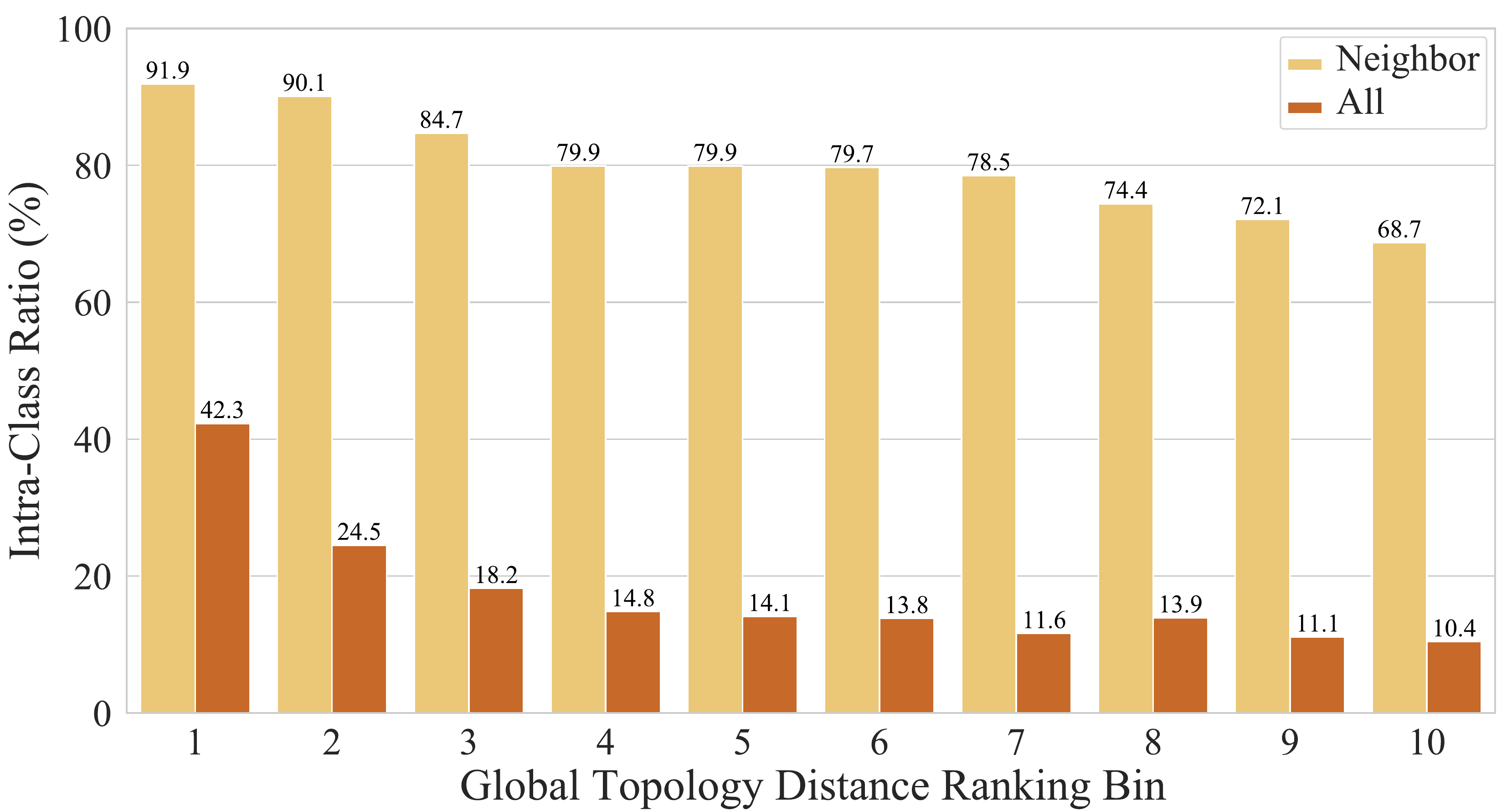}
\caption{The intra-class node-pair ratio across the global topology distance (Eq.~\ref{equation_gpr_kl}) ranking bin from the smallest to the largest (\textit{CORA} dataset). \textit{Neighbor} represents this ratio in all neighborhood node-pairs; \textit{All} represents this ratio in all possible node-pairs. }
\label{figure_intra_ratio}
\end{figure}

\subsection{TIFA-Contrastive Pair Sampling}
\label{section_sampling}
Contrastive pair sampling, especially the negative pair sampling, has a crucial impact on  CL~\cite{hn_sample,kdd_understanding_negative}. 
Different from existing works which simply regard all other nodes~\cite{gcl_nodeAug,gcl_grace} or labeled nodes from other class~\cite{gcl_cg3} as negative samples, we propose to sample contrastive pairs based on the relative distance between nodes.

We observe that (Figure~\ref{figure_intra_ratio}) nodes with similar annotated information distribution are more likely to share the same class label with each other, and vice versa. 
Therefore, our measurement for the node relative distance mainly depends on the difference of node LP result, which contains both the global topology information and annotated information.
Specifically, we normalize the each node's LP vector and then calculate the KL divergence between nodes as the global topology distance:
\begin{equation}
\bm{D}^g_{i,j}  = \mathrm{KL}(\bm{p}_i,\bm{p}_j), \quad \quad
\bm{p}_i = \mathrm{softmax}(\bm{Z}^{\ast}),
\label{equation_gpr_kl}
\end{equation}
where $\mathrm{softmax}(\cdot)$ function is adopted to transfer the original LP matrix $(\bm{Z}^{\ast})$ into a probability distribution over all the categories.
Besides, we supplement the relative distance measuring from the view of local topology distance and the node embedding distance.
For local topology distance $\bm{D}^l_{i,j}$, we use the minimum jump hop number as an indicator.
For node embedding distance $\bm{D}^e_{i,j}$, we calculate the cosine distance between the nodes in embedding $\bm{X}$.
The final node relative distance is calculated by the weighted sum of the global/local topology distance and the node embedding distance:
\begin{equation}
    \bm{D}_{i,j}= {\rm S} (\bm{D}^g_{i,j}) +  \lambda_1 {\rm S}(\bm{D}^l_{i,j}) +  \lambda_2 {\rm S}(\bm{D}^e_{i,j}), \label{relative_distance}
\end{equation}
where $\lambda_1,\lambda_2$ is the weight of the two supplement items; ${\rm S(\cdot)}$ represents the scale operation to transfer the original value to $[0,1]$. 
Then we construct positive and negative pairs for each node individually. For the $i$-th node (anchor node), the positive set $\bm{P}_i$ is composed of the closest nodes with the smallest relative distance; for the negative set $\bm{N}_i$, we propose to take the semi-difficult nodes~\cite{hn_metric_semi1} which are neither too far or too close to the anchor node. Given the personally ranked node list $\bm{R}_{i}$ sorted by $\bm{D}_{i}$ from the smallest to the largest (the $i$-th node itself is excluded), $\bm{P}_i$ and $\bm{N}_i$ are truncated from $\bm{R}_{i}$, respectively:
\begin{equation}
\bm{P}_i = \bm{R}_i[0:\mathrm{post}_{end}], \,
\bm{N}_i = \bm{R}_i[\mathrm{negt}
_{beg}:\mathrm{negt}_{end}]  
\label{equation_negt_beg} ,
\end{equation}
where $\mathrm{post}_{end}$ is the end index for positive set; $\mathrm{negt}_{beg}$ and $\mathrm{negt}_{end}$ are the begin and end indexes for negative set.

\section{Experiments}
In this section, we first give a brief introduction to the experiment datasets and settings.
Followingly, we show the effectiveness of the proposed TIFA-GCL method by comparing with both advanced GNN models and other GCL methods in both transductive and inductive settings.

\subsection{Datasets and Experimental Settings}
We conduct experiments on six widely-used graph datasets, namely paper citations networks~\cite{dataset_ccp} (\textit{CORA},\textit{CiteSeer} and \textit{Pubmed}; these three datasets are the mostly-used benchmark networks~\cite{semi-supervised,model_gcn,model_gat} in the SSNC studies), Amazon Co-purchase networks~\cite{graph_pitfall} (\textit{Photo} and \textit{Computers}) and the enormous \textit{OGB-Products} graph~\cite{OGB}. 
More details about the datasets, experiment settings and hyper-parameters can be found in Appendix C.

\subsection{Overall Results}
\label{section_overall_result}
The overall results of our proposed methods in transductive and inductive setting are shown in Table~\ref{table_baseline} and~\ref{table_products}, respectively.
For the transductive experiments, we set two lines of baselines to verify the effectiveness of our method:(1) Popular GNN models, including GCN~\cite{model_gcn}, GAT~\cite{model_gat}, GraphSAGE~\cite{model_sage} and GraphMix~\cite{model_graphmix}; (2) Existing GCL methods which apply CL to graph uniformly (\textbf{uniform-GCL}), including GRACE~\cite{gcl_grace}, NodeAug~\cite{gcl_nodeAug}, MultiView~\cite{gcl_multiview} and GraphCL~\cite{gcl_nipsaug}.
The transductive results are shown in Table~\ref{table_baseline}.
We can find that: 
(1) Methods with GCL modules (middle 4 models) generally outperform methods without GCL modules (top 4 models), since they can leverage the information from a large number of unsupervised nodes by comparing the perturbed representations.
(2) Our TIFA-GCL can further improve the GCL effect in SSNC than uniform-GCL methods by considering the received annotated information of different nodes.

For the inductive training, we conduct experiments to verify the effect of both TIFA-GCL framework and SAINT-TIFA subgraph sampling method. 
In Table~\ref{table_products}, we display results of different combinations of GNN aggregators, contrastive methods, and subgraph sampling methods (Full-batch,
Neighbor~\cite{model_sage}, SAINT-RW~\cite{GRAPHSAINT} and the proposed SAINT-TIFA).
We find that for both GAT and GraphSAGE\footnote{As shown in Table~\ref{table_products}, GCN performs worse than GAT and GraphSAGE, so we don't experiment on it. The same reason not to experiment with Neighbor Sampling in GCL.}, TIFA-GCL can bring larger improvement for GNN training compared to uniform-GCL\footnote{The uniform-GCL's performance here is based on our model's ablation version which replaces all the TIFA modules with the uniform ones from GRACE~\cite{gcl_grace}. The same below.}, which verifies the effectiveness of TIFA-GCL for inductive learning on enormous graphs.
Moreover, our SAINT-TIFA subgraph sampling method outperforms the SAINT-RW method for both GAT and GraphSAGE, proving that making nodes more likely to interact with high-affinity neighbors can further benefit inductive graph training. 

\begin{table}[t]
\centering

\setlength{\tabcolsep}{5pt}
\resizebox{.95\columnwidth}{!}
{
\begin{tabular}{@{}l|ccccc@{}}
\toprule
\textbf{Method}           & \textbf{\small CORA} & \textbf{\small CiteSeer} & \textbf{\small PubMed} & \textbf{\small Photo} & \textbf{\small Comp.} \\ \midrule
GCN      &     81.0\tiny$\pm$0.7           &     69.1\tiny$\pm$0.8              &   76.8\tiny$\pm$0.8             &    88.0\tiny$\pm$1.7              &      80.7\tiny$\pm$0.7             \\
GAT      &   79.9\tiny$\pm$2.1           &    69.6\tiny$\pm$1.4                &      77.2\tiny$\pm$1.0            &   88.7\tiny$\pm$1.9               &      77.9\tiny$\pm$2.1            \\
GraphSAGE     &    80.7\tiny$\pm$1.2          &       67.8\tiny$\pm$1.4             &      76.3\tiny$\pm$0.6           &        87.7\tiny$\pm$1.2         &    79.6\tiny$\pm$1.2              \\
GraphMix   &   81.5\tiny$\pm$0.9           &      69.3\tiny$\pm$0.4               &    78.0\tiny$\pm$1.3             &   89.5\tiny$\pm$1.7             &    81.2\tiny$\pm$2.7             \\ \midrule
GRACE      &  82.1\tiny$\pm$1.6              &     70.3\tiny$\pm$1.5               &     77.8\tiny$\pm$1.8            &    89.3\tiny$\pm$1.3             &   82.1\tiny$\pm$1.5               \\
NodeAug  &  82.7\tiny$\pm$1.3             &    71.1\tiny$\pm$1.2                &      78.4\tiny$\pm$0.7           &   89.7\tiny$\pm$1.8              &   83.0\tiny$\pm$1.9               \\
MultiView  &  81.8\tiny$\pm$0.8               &     71.5\tiny$\pm$1.7               &     77.9\tiny$\pm$0.1            &     88.9\tiny$\pm$1.5           &   82.8\tiny$\pm$2.0               \\
GraphCL  & 82.4\tiny$\pm$1.1              &      70.8\tiny$\pm$0.6              &       78.2\tiny$\pm$0.5          &    90.3\tiny$\pm$2.3             &  83.2\tiny$\pm$0.7                \\ \midrule
TIFA-GCL & \textbf{83.6}\tiny$\pm$0.8       &     \textbf{72.5}\tiny$\pm$0.6              &       \textbf{79.2}\tiny$\pm$0.6                  &    \textbf{91.0}\tiny$\pm$0.6             &   \textbf{83.6}\tiny$\pm$1.4               \\
\bottomrule
\end{tabular}}
\caption{Empirical results (classification accuracy\%) of graph transductive learning.
The top four methods are typical GNN models without a GCL mechanism, and the middle four methods are uniform-GCL methods.}
\label{table_baseline}
\end{table}

\subsection{Generalizability of TIFA-GCL}
\begin{figure*}[t]
\centering
\includegraphics[width=\linewidth]{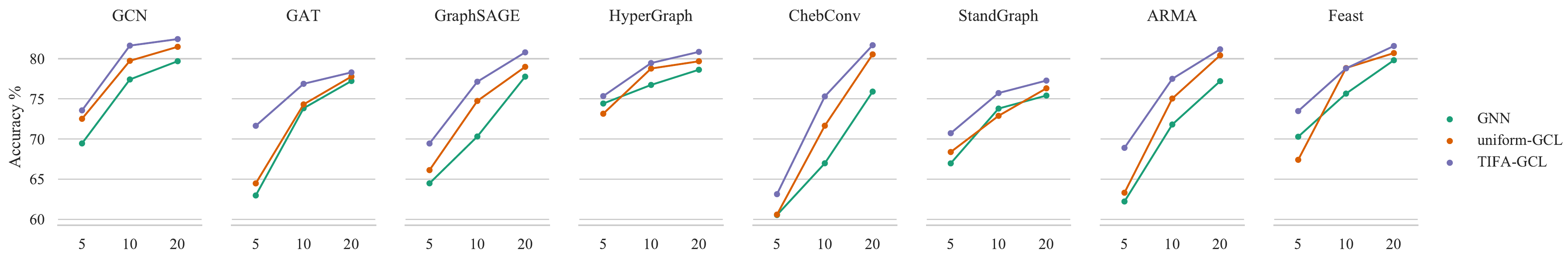}
\caption{The performance of TIFA-GCL on multiple base GNN backbones under different labeling size
~(5/10/20 labeled nodes in each category) on the \textit{CORA} dataset.} 
\label{figure_gnn}
\end{figure*}

Our TIFA-GCL method is model-agnostic and thus can be applied to any GNN encoders. To verify the generalizability of TIFA-GCL, we evaluate the performance of our TIFA-GCL with multiple GNN architectures and different labeling sizes.
As shown in Figure~\ref{figure_gnn}, we conduct experiments on 8 different GNN models: GCN~\cite{model_gcn}, GAT~\cite{model_gat},  GraphSAGE~\cite{model_sage}, HyperGraph~\cite{model_hyper_graph},  ChebConv~\cite{model_cheb}, StandGraph~\cite{model_stand_graph}, ARMA~\cite{model_arma} and Feast~\cite{model_feast}.
We find similar patterns in other datasets and thus only display the result of \textit{CORA} due to the space limit.
We find that, although these GNNs vary greatly in architectures, TIFA-GCL can consistently promote the GNN performance comparing to uniform-GCL under different labeling sizes, proving the effectiveness of TIFA-GCL in the broader SSNC scenarios.
We also find that the improvement of TIFA-GCL is more significant when the labeling size is small.
In such case, the annotated information is sparse and the importance of supplementary information from contrastive learning increases; thus our method is suitable for these scenes where few annotated data is available.
\begin{table}[t]
\centering
\setlength{\tabcolsep}{5pt}
\resizebox{.9\columnwidth}{!}
{
\begin{tabular}{@{}l|l|l|c@{}}
\toprule
 \textbf{Aggregator} & \textbf{\,\, Sampling}         & \textbf{Contrastive}   & \textbf{Test Acc(\%)}            \\ \midrule
MLP*       & N/A & No             &    
61.1\tiny$\pm$0.1\\
Node2vec*  & N/A & No            & 72.5\tiny$\pm$0.1 \\ \midrule
GCN*       & Full-batch     & No             & 
75.6\tiny$\pm$0.2 \\
GraphSAGE* & Full-batch     & No             & 78.5\tiny$\pm$0.1 \\
\midrule
\multirow{6}{*}{GraphSAGE}  & Neighbor  & No & 78.9\tiny$\pm$0.1\\
                            \cmidrule{2-4}
                            & \multirow{3}{*}{SAINT-RW}   & No & 79.3\tiny$\pm$0.2\\
                            &            & uniform-GCL & 79.8\tiny$\pm$0.4\\
                            &            & TIFA-GCL    &
                            80.1\tiny$\pm$0.4\\
                            \cmidrule{2-4}
                            & SAINT-TIFA & TIFA-GCL    & \textbf{80.7}\tiny$\pm$0.3\\
\midrule
\multirow{6}{*}{GAT}  & Neighbor  & No & 79.8\tiny$\pm$0.4\\
                            \cmidrule{2-4}
                            & \multirow{3}{*}{SAINT-RW}   & No & 80.0\tiny$\pm$0.4\\
                            &            & uniform-GCL & 81.1\tiny$\pm$0.1 \\
                            &            & TIFA-GCL    &
                            82.1\tiny$\pm$0.3\\
                            \cmidrule{2-4}
                            & SAINT-TIFA & TIFA-GCL    & \textbf{82.5}\tiny$\pm$0.5\\
\bottomrule
\end{tabular}}
\caption{Empirical results (classification accuracy\%) of inductive learning on \textit{OGB-Products} dataset. $\ast$ indicates that the result are taken from the OGB official results.}
\label{table_products}
\end{table}

\subsection{Ablation Study}
\label{section_ablation}
To verify the contributions of different modules in TIFA-GCL, we conduct ablation studies in this part.
Table~\ref{table_ablation} shows the performance of the model variants when removing the loss weight schedule from Section~\ref{section_schedule}, or replacing the graph perturbation from Section~\ref{section_graph_perb} and the pair sampling modules from Section~\ref{section_sampling} with the uniform one from GRACE~\cite{gcl_grace}.
From Table~\ref{table_ablation}, we find that when removing or replacing any one of three modules, the results of our model variants across three datasets all decrease, which verifies the effectiveness of them. 
We can also find that the TIFA-graph perturbation brings the largest improvement to the results; it is because that the strength of disturbance directly influences the GCL's effect and our TIFA-graph perturbation can provide intense contrastive information for subgraphs that needs supplementary information most.
Moreover, the results of these variants still outperform the the GCN and GRACE models, which demonstrates the overall advantage of the TIG-adaptive mechanisms in our approach.  

\begin{table}[t]
\centering
\setlength{\tabcolsep}{2pt}
\resizebox{.95\columnwidth}{!}
{
\begin{tabular}{@{}l|ccc@{}}
\toprule 
\textbf{Method}     & \textbf{CORA}           & \textbf{CiteSeer}      & \textbf{PubMed}   \\ \midrule
GCN      & 81.0\tiny$\pm$0.7       & 69.1\tiny$\pm$0.8      & 76.8\tiny$\pm$0.8 \\
GRACE      & 82.1{\tiny$\pm$1.6} 
& 70.3{\tiny$\pm$1.5} & 77.8{\tiny$\pm$1.8} \\ \midrule
Full TIFA-GCL        & 83.6\tiny$\pm$0.8       & 72.5\tiny$\pm$0.6      & 79.2\tiny$\pm$0.6 \\ 
$w/o$ TIFA-CL Schedule   & 82.7{\tiny$\pm$1.3} ($\downarrow 0.9$)  & 72.1{\tiny$\pm$0.6} ($\downarrow 0.4$)  & 78.7{\tiny$\pm$0.7} ($\downarrow 0.5$) \\ 
$r/p$  TIFA-Perturbation  &  82.8{\tiny$\pm$0.9} ($\downarrow 0.8$)  & 71.2{\tiny$\pm$1.2} ($\downarrow 1.3$)  & 78.5{\tiny$\pm$1.1} ($\downarrow 0.7$) \\
$r/p$  TIFA-Pair Sampling  &83.1{\tiny$\pm$1.3} ($\downarrow 0.5$)    &71.8{\tiny$\pm$0.5} ($\downarrow 0.7$)    &78.7{\tiny$\pm$0.7} ($\downarrow 0.5$) \\ 
\bottomrule
\end{tabular}}
\caption{Ablation results (classification accuracy \%) of the three TIFA modules in TIFA-GCL. $w/o$ denotes removing the module and $r/p$ denotes replacing the module with the uniform one in GRACE. }
\label{table_ablation}
\end{table}

\section{Analysis}
In this section, we provide in-depth analyses of our proposal, namely: 
(1) why TIFA-GCL can enhance GCL effect and 
(2) alleviate the over-smoothing problem in GNN training, 
(3) how to conduct graph-specific hard negative sampling 
and (4) when weighted neighbor sampling promoting inductive graph training.
Due to the limit of space, we present the discussion of the last three research questions in Appendix D.
\subsection{TIFA-GCL Enhance Under-represented Nodes}
\begin{figure}[t]
\center
\includegraphics[width=0.9\columnwidth]{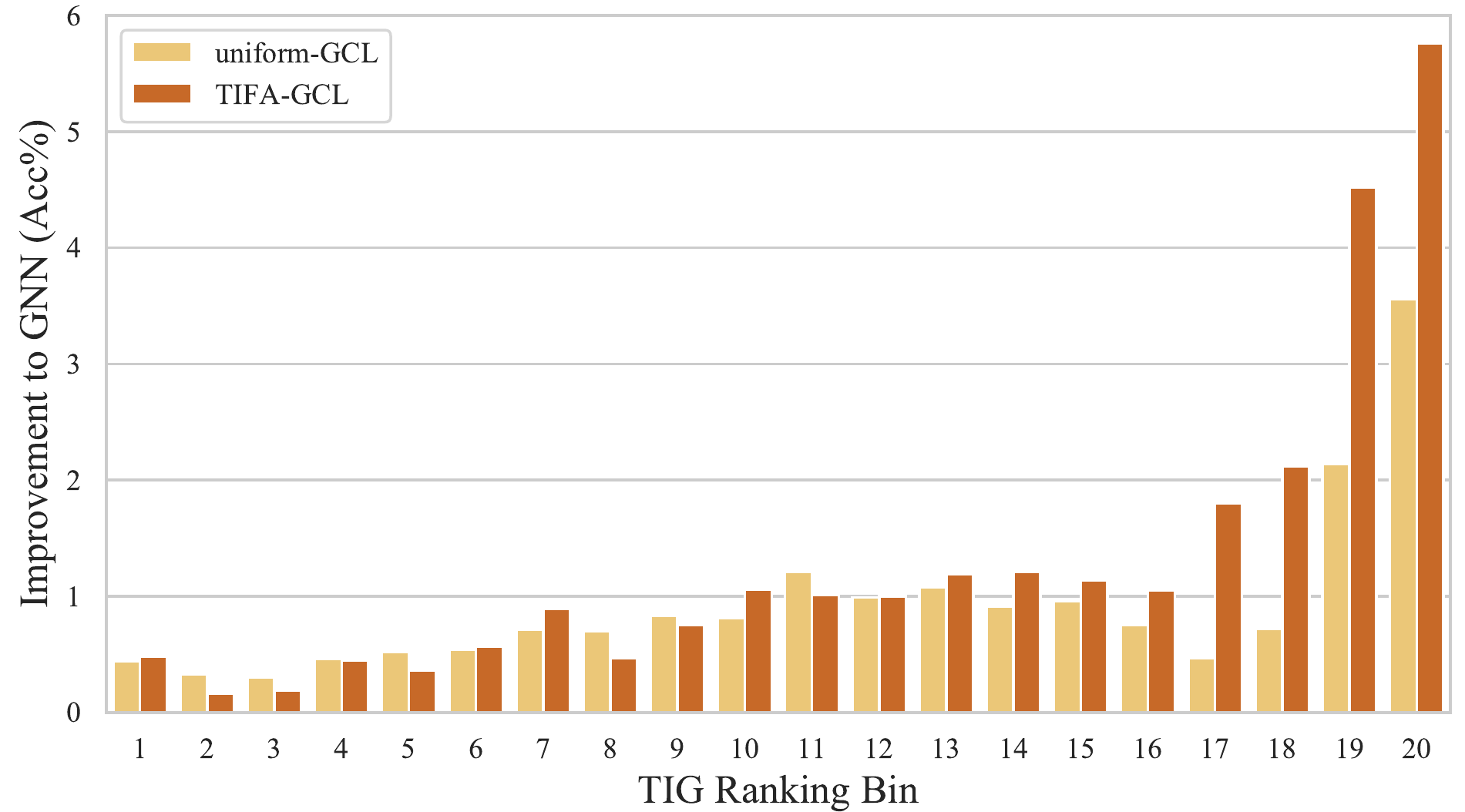}
\caption{The performance improvement gap across TIG ranking bins from the largest to the lowest.}
\label{figure_analysis2}
\end{figure}

To explore why our TIFA-GCL method can benefit SSNC more than uniform-GCL, we display the comparison between TIFA-GCL with uniform-GCL at each TIG ranking bin in Figure~\ref{figure_analysis2}. We can find that TIFA-GCL outperforms uniform-GCL by a large margin at nodes with small TIG values and achieves comparable performance at nodes with large TIG values. 
This indicates that our TIFA-GCL can effectively strengthen the GCL's effect for nodes that are under-represented in graph learning (small TIG value represents that node's received annotated information from topology is insufficient or confusing), thus achieving higher performance than existing uniform-GCL methods.

\section{Related Work}
The idea behind contrastive learning is to utilize self-supervision information among contrastive pairs, which are constructed by randomly perturbation. 
By regularizing the perturbed representations, e.g., pulling the distance between positive pairs closer while pushing negative pairs apart, CL learns high quality representations for both images~\cite{cl_SimCLR2,cl_fixmatch,cl_remixmatch} and texts~\cite{cl_mixtext,cl_coda,cl_capt}.

In graph field, \cite{gcl::dgi,gcl::hdgi} learn node representations via mutual information maximization, which can be treated as the early applications of the idea of CL in graph.
Recent GCL studies on SSNC mainly focus on three topics:
(1) {Devising graph-specific CL frameworks} by
combining GCL with different modules:
such as pre-training~\cite{gcl::gcc}, graph generating target~\cite{gcl_cg3}, implicit contrastive method~\cite{gcl::implicit}, and light-weighted regularization~\cite{gcl::reg}; 
(2) {Developing graph data augmentation methods}, such as graph diffusion~\cite{gcl_multiview}, random propagation~\cite{gcl_random_gnn}, subgraph aggregation~\cite{gcl::new_aug_method}, and generating from different GNN encoders~\cite{gcl_cg3};
and (3) {Analyzing important questions in GCL}, such as the augmentation method~\cite{gcl_nipsaug}, GCL robustness~\cite{gcl::robust}, sampling strategies~\cite{gcl::probing_negative}.
However, all these existing GCL methods ignore the imbalanced annotated information distribution and apply GCL uniformly to the whole graph. 
Our TIFA-GCL can solve this issue by considering the TIG of different nodes.

\section{Conclusion and Future Work}
In this work, we investigate contrastive learning in semi-supervised node classification. 
We point out that current GCL methods lack consideration of the uneven distribution of annotated information, which greatly affects the graph learning performance.
By analyzing the distributions of annotated information GCL contribution, we find that the benefit brought by GCL in node representation learning mainly comes from the nodes with less annotated information. 
Motivated by our findings, we systematically devise the TIFA-GCL framework. Extensive experiments in various scenarios demonstrate the effectiveness and generalizability of our method.

\section*{Acknowledgements}
We appreciate all the thoughtful and insightful suggestions from the anonymous reviews.
This work was supported in part by a Tencent Research Grant and National Natural Science Foundation of China (No. 62176002). Xu Sun is the corresponding author of this paper.

{\scriptsize
\bibliographystyle{named}
\bibliography{ijcai22}
}

\appendix

\section{Supplement to the Probing of GCL}
\subsection{Adjust LP from the Node Feature view.}
We propose that each unlabeled node owns its special preference for information from different classes, which is influenced by the affinity of node features.
Therefore, we adopt the similarity between the unlabeled node and the labeled node group's prototype to adjust the original LP vector $\bm{Z}_i$ from the node feature view:
\begin{flalign}
\bm{\mathcal{P}}_c &= \frac{1}{|\bm{{L}}_c|}\sum_{i\in\bm{{L}}_c}\bm{X}_i, \\
\bm{Z}_{i,c}^{\ast} &= \frac{1}{2}\bm{Z}_{i,c}\left(1+ \frac{\bm{X}_i \bm{\mathcal{P}}_c^{T}}{\|\bm{X}_i\| \|\bm{\mathcal{P}}_c\|}\right),
\end{flalign}
where $\bm{\mathcal{P}}_c $ is the prototype embedding of class $c$, calculated by averaging the embedding of all the labeled nodes in class $c$ \footnote{We try different similarity metrics and find the cosine method works best. 
}.
$\bm{Z}_{i,c}^{\ast}$ is applied in the analysis and experiment of this work.
Experiment results demonstrate that this modification from the node feature view can further improve the final performance of our method.

\subsection{Radial Decay Phenomenon in more Datasets}
\label{appendix::more_dataset}
\begin{figure*}[ht]
\centering
\includegraphics[width=\linewidth]{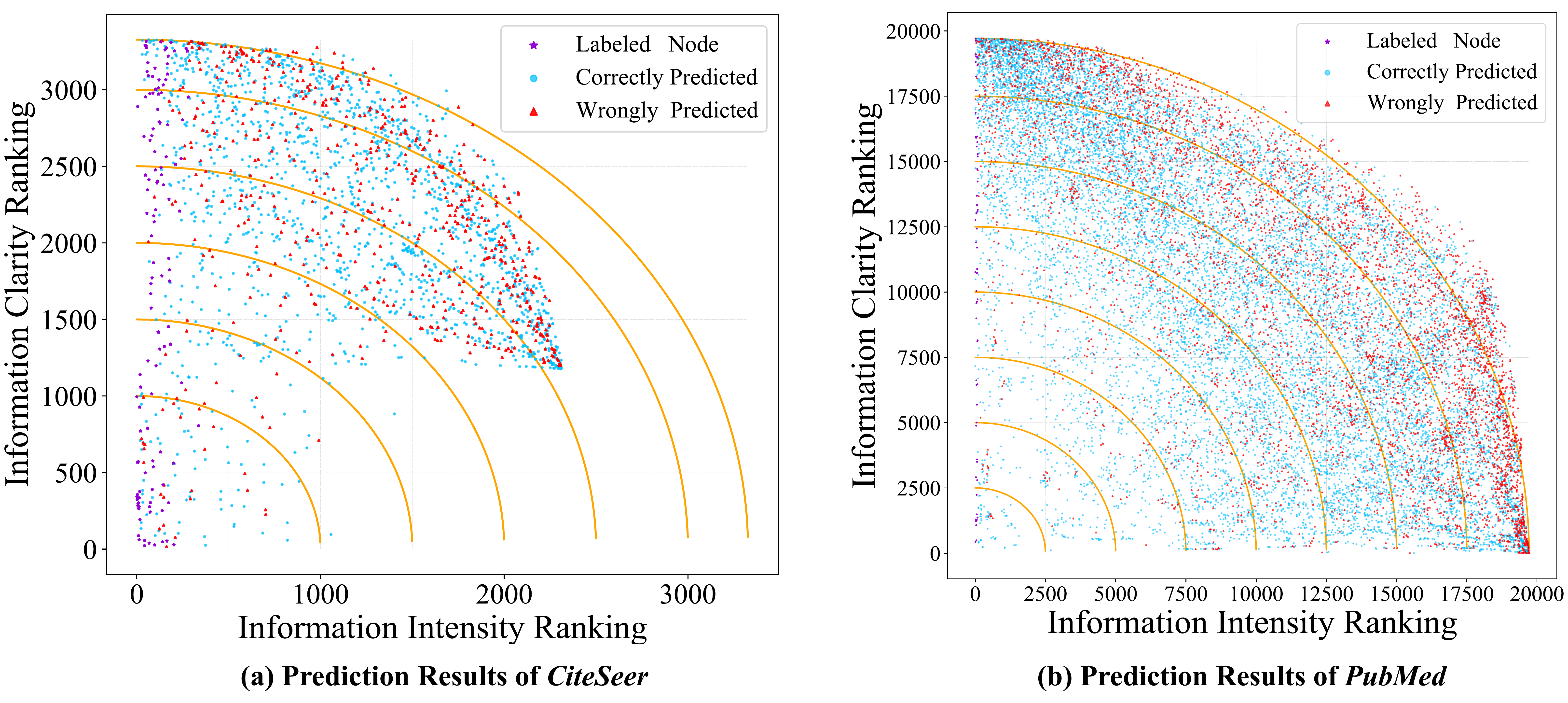}
\caption{
The \textit{RADIAL DECAY} phenomenon of SSNC in \textit{CiteSeer} and \textit{PubMed} datasets, which is in consistent with the phenomenon in \textit{CORA} dataset (Figure~1(a)). 
The white space in the lower right corner of the Subfigure~(a) is because we don't plot nodes that have no topology information.
\textit{CORA} and \textit{CiteSeer} datasets have some isolated nodes whose LP vector is filled with value $0$. \textit{PubMed} dataset has no isolated node.
}
\label{figure_appendix2}
\end{figure*}

We observe the same radial decay phenomenon in \textit{CiteSeer} and \textit{PubMed} datasets in Figure~\ref{figure_appendix2}, which is in consistent with the conclusion in the GCL working mechanism probing section.

\subsection{The GCL Effect across TIG Ranking Bins}
\label{appendix::bins}
\begin{figure*}[h!]
\centering
\includegraphics[width=\linewidth]{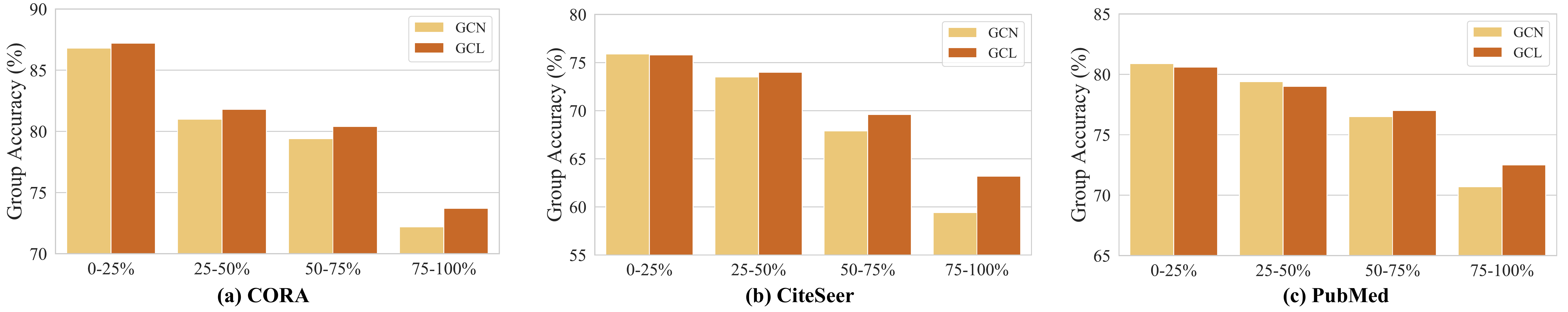}
\caption{The GCL effect across TIG ranking bins from the largest to the lowest (ranking 0\%-100\%). 
The GCL results here are taken from our implementation with GCN as aggregator.
The shared model architectures between GCN and GCL are strictly aligned.
We can obviously find that: 
(1) GNN performs worse on nodes with fewer TIG values, because the node's received task information from labeled nodes is insufficient or confuse;
and (2) GCL, in contrast, mainly improves GNN's learning ability on nodes with fewer TIG values.}
\label{figure_appendix1}
\end{figure*}

We show the GCL effect across TIG ranking bins in Figure~\ref{figure_appendix1}. We can conclude that the effect of GCL mainly comes from subgraphs with less annotated information.

\begin{algorithm}[h]
\centering
\footnotesize
\begin{algorithmic}[1]
\label{algorithm_saint_tifa}
\caption{SAINT-TIFA: Global Topology Distance--Weighted Subgraph Sampling}

\Require Graph $\mathcal{G} = (\mathcal{\bm{V}},\mathcal{\bm{E}}) $ with adjacency matrix $\bm{A}$; Sampling parameters: root nodes number $n$, random walk length $h$, sharpening coefficient $t$, the global topology distance matrix $\bm{D}^g$, error term $\varepsilon$.

\State $\mathcal{\bm{V}}_{root} \gets n$ root nodes uniformly sampled from $\mathcal{\bm{V}}$
\State $\mathcal{\bm{V}}_{s} \gets \mathcal{\bm{V}}_{root}$
\State $\bm{P}(i,j) \gets \frac{\bm{A}(i,j)}{\bm{D}^g(i,j) + \varepsilon}$
 \Comment{\textit{Make nodes with similar received annotated information more likely to be sampled; mask non-neighbor pairs.}}
\State $\bm{P}((i,j) \in \mathcal{\bm{E}}) = \frac{\mathrm{exp}(\bm{P}(i,j) \ast t)}{\sum_{k\in\bm{A}(i,:)}\mathrm{exp}(\bm{P}(i,k) \ast t)}$
 \Comment{\textit{Sharpen sampling probability distribution with} $t$.}
\For{$v \in \mathcal{\bm{V}}_{root}$:}
\State $u \gets v$
\For {$x = 1$ to $h$}
\State $u \gets $ Node sampled from $u's$ neighbors by $\bm{P}(u,:)$
\State  $\mathcal{\bm{V}}_{s} \gets \mathcal{\bm{V}}_{s} \cup \{u\} $
\EndFor
\EndFor
\State $\mathcal{G}_s \gets$ Node induced subgraph of $\mathcal{G}$ from $\mathcal{V}_s$
\Ensure $\mathcal{G}_s$
\end{algorithmic}
\end{algorithm}

\begin{algorithm}[h!]
\centering
\footnotesize
\begin{algorithmic}[1]
\label{algorithm::perturbation}
\caption{TIFA-Graph Perturbation with Negative Feedback}

\Require Graph $\mathcal{G} = (\mathcal{\bm{V}},\mathcal{\bm{E}}) $ with adjacency matrix $\bm{A}$; Node embedding dim $h$, CL weight $\bm{w}$, sharpen coefficient $t$, perturbation threshold $\sigma$,
maximum edge adding/removing number $n_{add}, n_{rmv}$,
feature mask rate $m$, probability decay hop range $d_h$ and probability decay ratio $d_r$.
\Statex
\Function {\textbf{PROBABILITY\_DECAY}}{$\bm{P}, \mathrm{i}$}
\Comment{\textit{Reduce the perturbation probability of subgraphs around the nodes that have been perturbed.}}
\State $\bm{P}[i] = 0$
\For{$x \in$ node $i's$ neighbors within $d_h$ hops}
\State $\bm{P}[x] = \bm{P}[x] * d_r$
\EndFor

\State $\bm{P}[i]  = \frac{\mathrm{exp}(\bm{P}[i])}{\sum_{k \in \mathcal{\bm{V}}} \mathrm{exp}(\bm{P}[i])} $
\State return $\bm{P} $
\EndFunction

\State
\State $\bm{X_p} \gets \bm{X}, \quad \bm{A_p} \gets \bm{A}$
\State $\bm{P}[i]  = \frac{\mathrm{exp}(w_i \ast t)}{\sum_{k \in \mathcal{\bm{V}}} \mathrm{exp}(w_k \ast t)} $
\Comment{\textit{Sharpen the selection probability distribution}}
\State $\mathrm{gap} \gets 0 $

\While{$\mathrm{gap} < \sigma$}
\State $i \gets$ sampled from $\mathcal{\bm{V}}$ according to probability $\bm{P}$
\State $\mathcal{\bm{V}}_{add} \gets n_{add}$ nodes uniformly sampled from $i's$ nonadjacent nodes  
\For{$x \in \mathcal{\bm{V}}_{add}$}
\State $\bm{A_p}[x][\mathrm{i}] = \bm{A_p}[\mathrm{i}][x] = 1$
\Comment{\textit{Randomly add edges}}
\EndFor

\State $\mathcal{\bm{V}}_{rmv} \gets n_{rmv}$ nodes uniformly sampled from $i's$ adjacent nodes  
\For{$x \in \mathcal{\bm{V}}_{rmv}$}
\State $\bm{A_p}[x][\mathrm{i}] = \bm{A_p}[\mathrm{i}][x] = 0$
\Comment{\textit{Randomly remove edges}}
\EndFor

\State $\mathbf{mask_{idx}} \gets (m \ast h)$ ids uniformly sampled from $1-h$
\For{$x \in \mathbf{mask_{idx}} $}
\State $\bm{X_p}[x] = 0$
\Comment{\textit{Randomly mask features.}}
\EndFor

\State $\bm{P} =$\textbf{PROBABILITY\_DECAY}$(\bm{P},\mathrm{i})$
\State $\mathrm{gap} = \sqrt{\sum_{i=1}^{|\mathcal{\bm{V}}|}\sum_{j=1}^{|\mathcal{\bm{V}}|}(\bm{A_p}[i][j] - \bm{A}[i][j])^{2}}$
\Comment{\textit{Calculating the Frobenius norm}}
\EndWhile
\Statex
\Ensure $\bm{X_p},\bm{A_p}$
\end{algorithmic}
\end{algorithm}

\section{Detailed Algorithms in Methodology}

\subsection{TIFA-Graph Perturbation Algorithm}
Data augmentation~(perturbation) is an essential component in contrastive learning. In existing studies, there are two general ideas to remedy this issue: randomly/heuristically choosing part of graph nodes to disturb~\cite{gcl_grace,gcl_adaAug} or augmenting each node in its parallel universe~\cite{gcl_nodeAug}.
However, neither of these two approaches take account of the differences of nodes' received task information from labeled nodes, which determine the node's requirements for supplementary information from CL.

Following existing studies~\cite{gcl_grace,gcl_nodeAug,gcl_nipsaug}, three kinds of augmentation operations are taken: (1) Randomly adding edges; (2) Randomly removing edges; and (3) Randomly masking node features.
We conduct sequential selections for nodes, and dynamically adjust the selecting probability by reducing the probability of subgraphs around the augmented node, which aims to remedy the excessive issue caused by the spreading of graph perturbation.
The algorithm stops once the graph disturbance reaches a preset threshold. The details of TIFA-graph perturbation is shown in Algorithm~2.

\subsection{TIFA-GCL for Inductive Graph Learning}
\label{section::saint_tifa}
Transductive GNN training methods~\cite{model_gcn,semi-supervised,model_gat} take the full graph as input, which can hardly be applied to scenarios where the graphs are too large to train in a full-batch manner.
Generally, there are two kinds of solutions to address this issue and conduct inductive training: (1) \cite{model_sage} propose the GraphSAGE method to train GNN on the entire graph and sample the neighbor sets during the training process, and (2) \cite{GRAPHSAINT} instead propose the GraphSAINT method to sample the subgraphs firstly and then train GNN on the subgraphs, which has proven more effective. Moreover, GraphSAINT-Random Walk (short as SAINT-RW) has proven especially effective in practice~\cite{GRAPHSAINT}.

Our TIFA-GCL method can be seamlessly applied to the inductive learning scenario with the SAINT-RW subgraph sampling method.
However, SAINT-RW conducts sampling uniformly from all neighbor nodes, which overlooks the difference in node affinity with different neighbors.
\cite{chen_smoothing} have proved that the working mechanism of GNN mainly lies in the interaction and smoothness of high-affinity node-pairs containing similar information; and in Figure 2 from the paper body, we find that neighbors with similar topology information distribution are more likely to share the same class label.
Therefore, we propose \textbf{SAINT-TIFA}, a weighted neighbor sampling strategy to make nodes more likely to interact with high-affinity neighbors, thus boosting inductive GNN training.
Specifically, we employ the global topology distance from Eq.12 as an indicator to adjust the sampling probability for different neighbors.

The SAINT-TIFA subgraph sampling algorithm is shown in Algorithm~1.
We adjust the preference to high-affinity neighbours with the sharpening coefficient $t$.
The larger the value of $t$, the more likely the SAINT-TIFA algorithm will walk to neighbors with similar annotated information distribution.
The SAINT-TIFA sampling method degenerates to the standard SAINT-RW method when $t$ is zero. 

\section{Detailed Experiment Settings for Reproducibility}
\subsection{Linux Server}
We run all the experiments on a Linux server, some important information is listed:
\begin{itemize}
    \item CPU: Intel(R) Xeon(R) Silver 4210 CPU @ 2.20GHz $\times$ 40
    \item GPU: NVIDIA GeForce RTX2080TI-11GB $\times$ 8
    \item RAM: 125GB
    \item cuda: 10.1.243
    \item cudann: 7.6.5.32-1+cuda10.1
\end{itemize}

\subsection{Python Package}
We implement all deep learning methods based on Python3.7. The versions of some important packages are listed:
\begin{itemize}
    \item torch: 1.6.0+cu101
    \item torch-geometric: 1.6.1
    \item torch-cluster:1.5.7
    \item torch-sparse: 0.6.7
    \item ogb: 1.2.4
    \item scikit-learn: 0.23.2
    \item numpy:1.19.1
    \item scipy:1.5.2
\end{itemize}

\subsection{Hyper-parameters Selection for TIFA-GCL Framework}
In this part, we introduce the selection range for the model-agnostic hyper-parameters in TIFA-GCL training framework. All the hyper-parameters are tuned on the validation set. Selection list with bold fonts indicates that model usually perform best with these values. Otherwise, it means that the parameters are sensitive to different graph datasets and GNN models.
\begin{itemize}
    \item Random walk teleport probability $\alpha$ in Eq.2:\\
    $\{0.05,0.1,\bm{0.15},0.2,0.25\}$,
    \item Weight of topology information clarity comparing to intensity $\lambda$ in Eq.7:\\
    $\{0,0.05,\bm{0.1},0.15,0.2,0.5\}$,
    \item Weight of CL loss with negative pairs comparing to positive pairs $\mu_1$ in Eq.11:\\
    $\{0,0.1,0.25,\bm{0.33},\bm{0.5},1\}$,
    \item Weight of CL loss with node pairs comparing to self-consistence loss $\mu_2$ in Eq.12:\\
    $\{0,0.05,0.1,0.15,0.2,0.25\}$,
    \item Minimum value of CL loss weight comparing to CE loss in Eq.13:\\
    $\{0.5,0.75,\bm{1},1.5,2,2.5,3,3.5\}$,
    \item Maximum value of CL loss weight comparing to CE loss in Eq.13:\\
    $\{1,1.5,\bm{2},2.5,3,3.5,4,4.5,5\}$,
    \item Weight of local topology distance in relative distance measuring comparing to global topology distance $\lambda_1$:\\
    $\{0,0.25,\bm{0.5},0.75,1,1.25,1.5\}$,
    \item Weight of node embedding distance in relative distance measuring comparing to global topology distance $\lambda_2$:\\
    $\{0,0.25,0.5,\bm{0.75},1,1.25,1.5\}$.
\end{itemize}

\subsection{Hyper-parameters Selection for GNN Aggregator}
In this part, we introduce the selection range for the model-specific hyper-parameters for GNN aggregators.
For all comparable GNN encoders, we stack GNN layers with $\mathrm{ReLU}$~\cite{relu} activation function. 
We use the Adam optimizer~\cite{kingma2014adam} and early stopping strategy in training.
\begin{itemize}
    \item Hidden Size:$\{16,32,\bm{64},\bm{128},256\}$ 64 for transductive learning and 128 for inductive learning,
    \item GNN Layer:$\{\bm{2},\bm{3},4,5\}$ 2 for transductive learning and 3 for inductive learning,
    \item Learning Rate:$\{0.005,\bm{0.0075},0.01,0.0125\}$,
    \item Dropout Probability:
    $\{0.2,0.3,0.4,\bm{0.5},0.6 \}$.
\end{itemize}
Besides, the maximum training epoch is set to $200$; training will last for at least $30$ epochs; after that, training will early stop if there is no improvement on the validation set for $20$ epochs. We adopt $L_2$ regularization with weight $0.005$. The learning rate begins to decay after $30$ epochs with decay coefficient $0.95$.

\subsection{More Information about Experiment Datasets}
The statistical information of the used datasets in the experiments is shown in Table~\ref{table_dataset}.

For all the datasets except for \textit{OGB-Products}, we use the standard transductive semi-supervised node classification setting~\cite{semi-supervised}, i.e., each category has $20$ nodes for training and $30$ nodes for validation, and all the rest nodes are used for testing.
Following~\cite{graph_pitfall} and ~\cite{method_fishergcn}, we run 20 random dataset splittings for these five datasets to relieve the random error caused by dataset splitting. Besides, we repeat the experiment 5 times for each dataset splitting.  
For the inductive training dataset \textit{OGB-Products}, we use the official~\cite{OGB} dataset splitting and repeat the experiment 10 times.
The backbone GNN aggregator among all methods is GCN (unless otherwise stated).
In each group of contrast experiments, the model architectures shared between the baselines and proposed methods are strictly aligned, to clearly show the pure effect brought by our method.
\begin{table}[t]
\centering
\small
\sisetup{detect-all,mode=text,group-separator={,},group-minimum-digits=3,input-decimal-markers=.}
\resizebox{.9\columnwidth}{!}
{
\begin{tabular}{@{}l|rrrrrc@{}}
\toprule
\bf Dataset     & \textbf{\#Node} & \textbf{\#Edge} & \textbf{\#Labeled} & \textbf{\#Class}  &\textbf{Train}  \\\midrule
CORA           & 2,708          & 5,429         & 140      & 7               &   Trans.                 \\
CiteSeer       & 3,327          & 4,732         & 120      & 6             &   Trans.             \\
PubMed         & 19,717         & 44,338        & 60       & 3                 &   Trans.                    \\
Photo          & 7,487          & 119,043       & 160      & 8                   &   Trans.                    \\
Comp.      & 13,381         & 245,778       & 200      & 10                  &   Trans.                    \\
Products   & 2,449,029      & 123,718,280   & 196,615  & 47                  &   Induc. \\
\bottomrule
\end{tabular}}
\caption{Statistical information about datasets.
 \textit{Trans.} indicates transductive learning and \textit{Induc.} indicates inductive learning. \textit{Comp.} is the short for the \textit{Computers} dataset.}
\label{table_dataset}
\end{table}

\begin{table}[t]
\centering
\resizebox{.95\columnwidth}{!}
{
\begin{tabular}{@{}ccccccc@{}}
\toprule 
\multirow{2}{*}[-3pt]{\bf Layer}       & \multicolumn{3}{c}{\textbf{Accuracy\%}~($\uparrow$)}  & \multicolumn{3}{c}{\textbf{MADGap}~($\uparrow$)} \\ \cmidrule(lr){2-4} \cmidrule(lr){5-7}
 & GCN    & GCL   & TIFA-GCL   & GCN    & GCL    & TIFA-GCL     \\\midrule
2     &  79.3\tiny$\pm$1.7       &  81.6\tiny$\pm$1.8     &   83.6\tiny$\pm$0.8      &  0.82\tiny$\pm$0.02      &  0.85\tiny$\pm$0.02      &  0.87\tiny$\pm$0.01       \\
4    &  76.2\tiny$\pm$1.2       &  81.3\tiny$\pm$1.6     &   82.7\tiny$\pm$1.4      &  0.77\tiny$\pm$0.04      &  0.78\tiny$\pm$0.02      &  0.78\tiny$\pm$0.01        \\
6  &  62.0\tiny$\pm$9.9       &  77.0\tiny$\pm$2.5     &   79.5\tiny$\pm$1.5      &  0.66\tiny$\pm$0.10      &  0.76\tiny$\pm$0.02      &  0.77\tiny$\pm$0.02        \\
8    &  \,43.5\tiny$\pm$10.3      &  68.5\tiny$\pm$8.2     &   75.8\tiny$\pm$4.7      &  0.62\tiny$\pm$0.10      &  0.65\tiny$\pm$0.08      &  0.72\tiny$\pm$0.05        \\
\bottomrule      
\end{tabular}}
\caption{Analysis of the over-smoothing phenomenon in deep GNNs. MADGap measures the GNN over-smoothing degree. $\uparrow$ means the bigger the better. We can find that TIFA-GCL can effectively relieve the over-smoothing phenomenon and improve model performance for deep GNNs.}
\label{table_smoothing}
\end{table}

\begin{figure}[t]
\centering
\includegraphics[width=0.9\columnwidth]{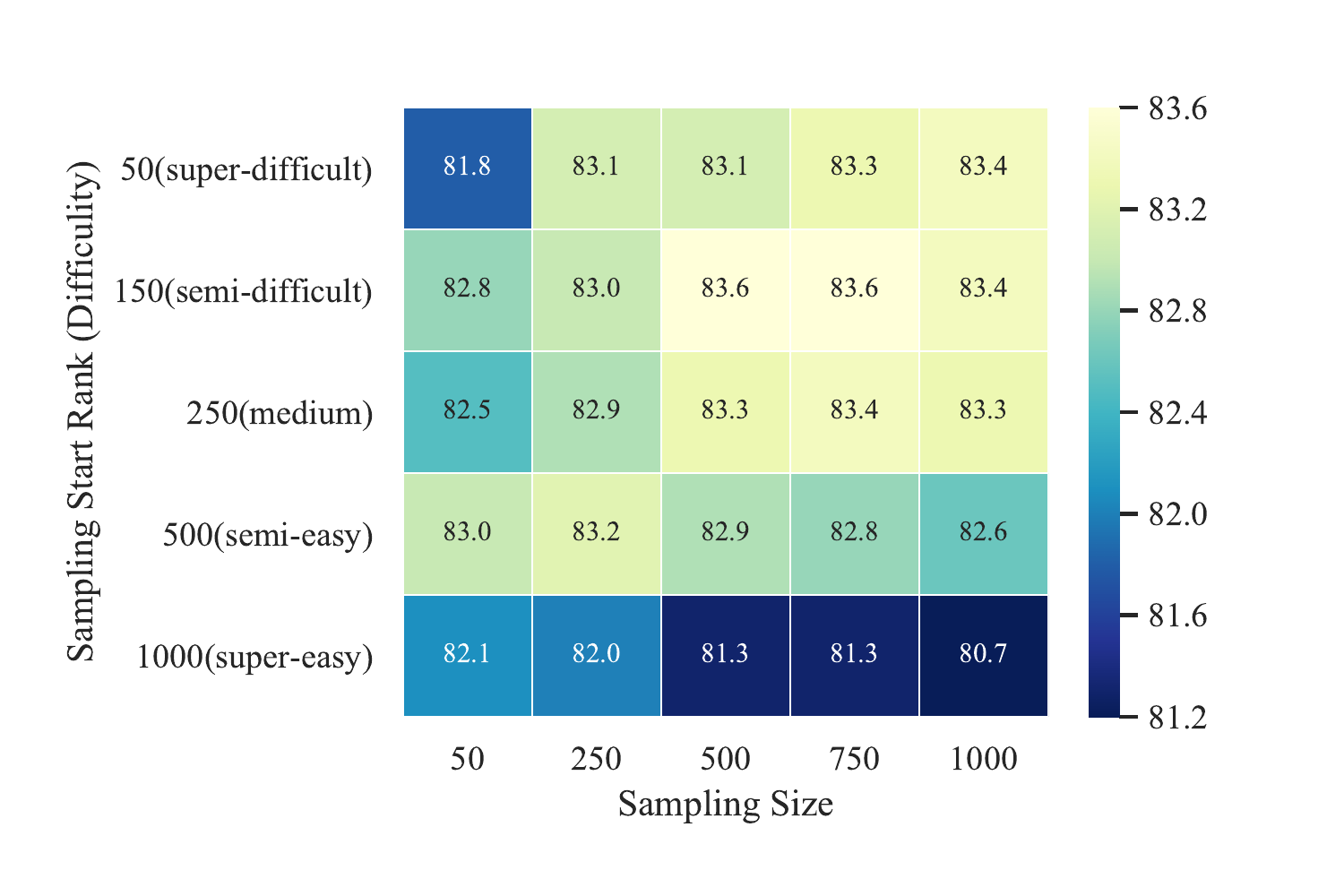}
\caption{Heatmap of TIFA-GCL's performance under different negative pair sampling settings. }
\label{figure_heatmap}
\end{figure}

\section{Supplement to Analysis}

\subsection{TIFA-GCL Effectively Alleviates GNN Over-smoothing Issue}

Over-smoothing is a common issue in GNN~\cite{PairNorm,analysis_smoothing}, especially for GNNs with multiple layers stacked.
In Table~\ref{table_smoothing}, we display  MADGap~\cite{chen_smoothing,gcl_random_gnn}, an indicator for GNN over-smoothness, of \textit{CORA} dataset trained by GCN models with various layers. 
We find that TIFA-GCL can effectively relieve the over-smoothing problem, thus promoting the GNN performance even for deep GNNs. 
We analyze the reason lies in that, the interactions of graph nodes make nodes similar with each other~\cite{chen_smoothing}, while vanilla GNN training lacks the mechanism for pulling representations of nodes apart. 
GCL can relieve this issue via self-consistency or pair-wise regularization; therefore the contrastive pairs are crucial in this process. 
TIFA-GCL sample contrastive pairs for each node individually by measuring relative distance, thus it can sample the more appropriate contrastive pairs and alleviate the over-smoothing phenomenon more effectively than uniform-GCL.

\subsection{Trade-off between Sampling Size and Difficulty of Negative Contrastive Pair}
Contrastive pair sampling has proven essential for the  CL performance~\cite{cl_moco1,cl_SimCLR2}. 
In Eq.17, we sample contrastive pairs by measuring the relative distance from the anchor node to all the other nodes.
Empirical results show that the closest nodes work best as the positive samples, and for the negative pair sampling, we further study the influence of sampling size and difficulty in Figure~\ref{figure_heatmap}. 
We observe that model can hardly be trained well by too easy or too difficult negative pairs, while the semi-difficult nodes serve best as negative samples.
This phenomenon is in consistent with CL studies~\cite{hn_metric_semi1,hn_sample} in scenes other than graph and proves the effectiveness of our relative distance measuring (Eq.16).
Besides, we notice that increasing the negative sampling size can effectively improve CL performance, because more contrastive pairs can provide more reliable target~\cite{cl_moco1}. 
However, a large sampling size will naturally reduce the difficulty of the selected pairs, telling that there is a trade-off between sampling size and difficulty of contrastive pairs.

\subsection{When Weighted Neighbor Sampling Promotes Inductive Graph Training}
\label{section_analysis_TIFA-SAINT}
We propose the SAINT-TIFA method to sample high-affinity neighbors instead of the random sampling method~\cite{GRAPHSAINT}.
In this part, we further study the influence of the preference to high-affinity neighbors.
In Table~\ref{table_ana4}, we sharpen the distribution of sampling probability under different random walk steps.  
We can find that, when we concentrate the sampling probability to the high-affinity neighbors, the model performance can be effectively improved, especially when the walking length is large.
However, it is worth noticing that when the sharpening coefficient is relatively large or the random walk length is small, the model performance declines obviously. We speculate the reason lies in that a GNN model needs to learn some negative examples for training to avoid the trivial solution. 
We conclude from this experiment that, when the walking length is small, dispersing the sampling probability distribution is the optimal;
while when the walking length increases, a larger sharpening coefficient can be taken to improve the quality of sampled subgraphs, thus promoting the inductive graph learning result.

\begin{table}[t]
\centering
\resizebox{.8\columnwidth}{!}
{
\begin{tabular}{@{}cc|llll@{}}
\toprule
\textbf{\small $t$} & \textbf{\small STD.} & \textbf{WL=2} & \textbf{WL=3} & \textbf{WL=4} & \textbf{WL=5} \\\midrule
0.00        & 0.000        & 81.6\tiny$\pm$0.8     & 82.1\tiny$\pm$0.3        & 80.4\tiny$\pm$0.9     & 80.4\tiny$\pm$0.9      \\
0.25        & 0.010        & 81.2\tiny$\pm$1.6     & \textbf{82.5}\tiny$\pm$0.5        & 81.2\tiny$\pm$1.2     & 80.2\tiny$\pm$2.4     \\
0.50        & 0.022        & 79.6\tiny$\pm$1.4     & 81.4\tiny$\pm$1.2        & 81.9\tiny$\pm$1.3     & 80.8\tiny$\pm$3.0     \\
1.00        & 0.032        & 79.9\tiny$\pm$1.9     & 80.7\tiny$\pm$1.6        & 80.7\tiny$\pm$1.4     & 80.9\tiny$\pm$2.1     \\
2.00        & 0.040        & 79.3\tiny$\pm$2.7     & 79.3\tiny$\pm$0.9        & 79.4\tiny$\pm$1.2     & 79.7\tiny$\pm$2.1     \\    
\bottomrule
\end{tabular}}
\caption{Analysis of the preference to high-affinity neighbors in SAINT-TIFA. \textbf{$t$} is the sharpening coefficient in Algorithm~1.
\textit{STD.} indicates the standard deviation for the sampling probabilities.
\textit{WL} is the length of walking steps.
}
\label{table_ana4}
\end{table}

\end{document}